# Predicting erectile dysfunction after treatment for localized prostate cancer

Hajar Hasannejadasl[1], Cheryl Roumen[1], Henk van der Poel[2], Ben Vanneste[3], Joep van Roermund[4], Katja Aben[5,6], Petros Kalendralis[1], Biche Osong[1], Lambertus Kiemeney[5], Inge Van Oort[7], Renee Verwey[8], Laura Hochstenbach[8], Esther J. Bloemen- van Gurp[8,9], Andre Dekker[1], Rianne R.R. Fijten[1]

1. Department of Radiation Oncology (Maastro), GROW School for Oncology, Maastricht University Medical Centre+, 6229 ET Maastricht, The Netherlands
2. Department of Urology, Netherlands Cancer Institute, Amsterdam, The Netherlands
3. Maastro Clinic, Maastricht, The Netherlands
4. Department of Urology, Maastricht University Medical Center+, The Netherlands
5. Department of Research & Development, Netherlands Comprehensive Cancer Organization, Utrecht, The Netherlands
6. Institute for Health Sciences, Radboud university medical centre, Nijmegen, The Netherlands
7. Department of Urology, Radboud university Medical Center, Nijmegen, The Netherlands
8. Zuyd University of Applied Sciences, Heerlen, The Netherlands
9. Fontys University of Applied Sciences, Eindhoven, The Netherlands

## Abstract

While the 10-year survival rate for localized prostate cancer patients is very good (>98%), side effects of treatment may limit quality of life significantly. Erectile dysfunction (ED) is a common burden associated with increasing age as well as prostate cancer treatment. Although many studies have investigated the factors affecting erectile dysfunction (ED) after prostate cancer treatment, only limited studies have investigated whether ED can be predicted before the start of treatment. The advent of machine learning (ML) based prediction tools in oncology offers a promising approach to improve accuracy of prediction and quality of care. Predicting ED may help aid shared decision making by making the advantages and disadvantages of certain treatments clear, so that a tailored treatment for an individual patient can be chosen. This study aimed to predict ED at 1-year and 2-year post-diagnosis based on patient demographics, clinical data and patient-reported outcomes (PROMs) measured at diagnosis. We used a subset of the ProZIB dataset collected by the Netherlands Comprehensive Cancer Organization (Integraal Kankercentrum Nederland; IKNL) that contained information on 964 localized prostate cancer cases from 69 Dutch hospitals for model training and external validation. Two models were generated using a logistic regression algorithm coupled with Recursive Feature Elimination (RFE). The first predicted ED 1 year post-diagnosis and required 10 pre-treatment variables; the second predicted ED 2 years post-diagnosis with the same number of pre-treatment variables. The validation AUCs were 0.84 and 0.81 for 1 year and 2 years post-diagnosis respectively. In conclusion, we successfully developed and validated two models that predicted ED in patients with localized prostate cancer. These models will allow physicians and patients alike to make informed evidence-based decisions about the most suitable treatment with quality of life in mind.

# Introduction

In recent years, a transition is taking place in health care to more personalised and patient-centered care by implementing interventions like data-driven prediction modelling and shared decision making (1). In particular, artificial intelligence (AI) is attracting a lot of attention in healthcare because of its ability to extract and condense patient information from multi-source data. As a result, and if implemented correctly, this new technology could lead to increased efficiency and ultimately optimal use of available resources (2). For example, it could allow selection of the optimal treatment for an individual based on their patient characteristics, genetics, environmental factors and their lifestyle (3–5).

Of particular interest are patients for which the treatment methods are not superior to each other, and thus consideration of patients' preferences and values plays an important role (6). In oncology, this is often the case for patients with localized prostate cancer. Globally, prostate cancer is the second most common cancer among men (7,8). In the Netherlands, more than 12,000 men are diagnosed with prostate cancer every year, which makes it the most common cancer among men (9). Diagnosis and treatment of prostate cancer not only affects the patient's quality of life (QoL) in the short term.  Post-treatment QoL studies also indicate worsening of the patient's sexual, urinary and bowel problems in the long term (10). As a result, patients newly diagnosed with localized prostate cancer and their doctors face the challenge of choosing the appropriate treatment option from the main options including radical prostatectomy (RP), brachytherapy (BT), external beam radiotherapy (EBRT), and active surveillance (AS) (11). While the choice does not influence their survival, their long-term QoL could be drastically different if one choice is made vs. another. The most common side effect that affects QoL in these patients is erectile dysfunction (ED), a condition in which the patient is unable to get or reliably hold an erection (12–14). ED generally appears several months after treatment and affects many patients in the long term with over 80% reporting ED up to 42 months after diagnosis (14,15). The rate of ED largely depends on the treatment given; one year after treatment up to 24% of patients suffer from ED after BT, 27% after active surveillance, 45% after EBRT and 66-87% after surgery (nerve-sparing or non nerve-sparing) (16,17) Also, the prevalence of ED shows a age-related increase (18). In addition, ED development is also subject to individual characteristics of the patient, such as age (19), ethnicity (20), chronic kidney disorder (21), hyperlipidemia (22), diabetes and cardiovascular disease (23,24). A patient's lifestyle may also influence the chance of developing ED. For instance, smoking (25,26), diet (27) and alcohol use may influence ED development (28).

Despite the influence of individual characteristics of a patient on the development of post-treatment ED, only generic population-based ED statistics are currently provided to patients in clinical practice.

In this study we employed a dataset generated by the ProZIB study (ProstaatkankerZorg In Beeld or "Insight into Prostate Cancer Care") to develop and externally validate a personalised predictor of post-treatment ED based on patient characteristics. The ProZIB study was initiated as a collaboration between the Netherlands Comprehensive Cancer Organisation, all medical associations involved in prostate cancer care, and patient advocacy groups (17).

# Methods

## Dataset

A nationwide prospective study aimed at the evaluation of prostate cancer care in the Netherlands was performed in previous years, resulting in the ProZIB dataset (17). This study was embedded in the Netherlands Cancer Registry which is hosted by the Netherlands Comprehensive Cancer Organisation. A subset of this ProZIB dataset was provided for the current study and contains clinical data and patient reported outcome measures (PROMs) based on the EPIC26 questionnaire (19) at diagnosis and at 12 and 24 months after diagnosis. All patients with localized or locally advanced prostate cancer, but without regional lymph node involvement or distant metastasis (cT1-3, N0, M0) were included. Further details about the larger dataset are described in the article by Vernooij et al. (17).

At diagnosis the following data elements were available: (i) PROMs data from the EPIC26 questionnaire; (ii) tumor characteristics , such as tumor staging, PSA at diagnosis and ISUP (International Society of Urological Pathology) Gleason grade group; (iii) patient characteristics, such as age, height, weight, smoking status, comorbidities, and information regarding treatment. Next to the baseline PROMs, PROMs 12 and 24 months after diagnosis were available. In addition, the dataset also contained information about the hospital at which the diagnosis and/or treatment took place.

## Treatment categories

Various (combinations of) treatments were reported in the dataset. These were combined into 4 treatment categories that are generally chosen after diagnosis:
1. RP with or without lymph node dissection and with/without hormone therapy.
2. EBRT alone, with lymph node dissection, or with lymph node dissection and hormone therapy.
3. BT alone or combined with hormone therapy.
4. No active therapy (NAT) including AS as well as watchful waiting (WW).

Fifteen patients were excluded for not receiving any of the treatment options or receiving a combination of treatments. Hormone therapy was added as a separate variable alongside the selected treatment categories of interest.

This resulted in a dataset containing a total of 949 patients.

## Outcome

The outcome predicted in these prediction models is the frequency of erections, question 10 in the EPIC-26 questionnaire (figure 1).

**10. How would you describe the FREQUENCY of your erections during the last 4 weeks?**

I NEVER had an erection when I wanted one................................. 1
I had an erection LESS THAN HALF the time I wanted one............. 2
I had an erection ABOUT HALF the time I wanted one ................... 3     (Circle one number)
I had an erection MORE THAN HALF the time I wanted one............ 4
I had an erection WHENEVER I wanted one..................................... 5

Figure 1: the description of the question in the EPIC-26 questionnaire that describes the side effect of erection reliability.

To classify between patients that could never have an erection and those that still could, we performed a binary transformation of the outcome. This binary transformation combined patients that answered number 1 (never) into one group and those that answered 2-5 in the other group.

## Data preprocessing

### Missing data

Where possible, missing values were removed. First, missing values were removed on a variable level, where variables with high portions of missing values were removed. Then missing values were removed on a patient level, i.e. patients with more missing values than the 95th percentile of the distribution of missing values were removed. Since more values were missing in the EPIC-26 questionnaire data at the 2-year time point than the 1-year time point, two separate datasets were created to achieve the highest possible amount of data for both time points.

### Batch effects

Principal Component Analysis (PCA) (29) was used to investigate potential batch effects between groups of patients. Batch effects are potentially confounding effects present in the data that are correlated to the outcome that is to be predicted, but not of biological or medical origin. In our analyses, PCA was used to test whether the following input variables caused any batch effects in our dataset: the hospital of diagnosis, age, tumor staging, PSA level, the treatments given, the presence of diabetes, the presence of cardiovascular disease, alcohol use, and smoking status.

### Training and test set

The original dataset contained data from patients diagnosed and treated in 69 hospitals in the Netherlands. According to the TRIPOD statement (30), independent statistical validation is essential to guarantee reproducibility of prediction models in a wider population. It classifies external validation into two types; type 3 and 4. Type 3 TRIPOD validation refers to external validation with a dataset that is gathered at a different location or time. We created a type 3 validation dataset that differed based on location based on the following:

> *75% of the total number of patients were selected for the training set and 25% for the test set.*
> *AND*
> *75% of all hospitals in the dataset were selected for the training set and 25% for the test set, i.e. all patients from one hospital were selected for either the training set or the test set, but no hospital occured in both sets.*

with the following protocol:
1. Hospitals were sorted based on the number of patients diagnosed at that specific hospital.
2. The hospital with the largest number of patients was selected for the training set and the second-largest hospital was selected for the test set.
3. Each subsequently-largest hospital was then placed in either the training or test set based on the distribution of patients in the training set vs. test set (75% vs. 25% +- 5%). In other words, if the training set contained more than 80% of the selected patients, the next group of patients from one hospital was placed in the test set. If the test set exceeded 30% of the entire dataset, the next group of patients was added to the training set and so on.

## Data analysis

### Univariate analyses

Before the statistical modelling, we investigated which demographic variables were statistically significant between patients that could no longer get an erection and those that still could based on a Student's t-test in the case of a normal distribution or Wilcoxon rank test in the case of a non-normal distribution. In addition, a false discovery rate correction (FDR) (31) was applied using the R "fdrtool" package (32) to generate q-values and reduce the chances of reporting false positives. A similar approach was taken to investigate potential significant differences between the training and test set for the following variables: (i) age at diagnosis; (ii) clinical tumor stages; (iii) PSA at diagnosis; (iv) ISUP grade group.

### Multivariate analyses

To predict the ED outcome at 1 and 2 years post-diagnosis, we used a bootstrapped binomial logistic regression coupled with Recursive Feature Elimination from the "glmnet" (33) and "caret" (34) packages in the R programming language (version 3.6.3). After the development of the prediction models, Receiver Operating Curves (ROCs) were generated using the R "pROC" package (35) and sensitivity, specificity and overall accuracy were calculated. In addition, we created a nomogram of both logistic regression models with the "rms" package in R (36).

### Calibration

Despite the importance of discrimination in classifying patients, the degree of quantitative reliability of the model for the observed results is crucial in clinical decision-making. In well-calibrated models, the predicted results are consistent with the results of the study population. The results show that calibrated models are more helpful in practice even with lower AUC (37). For the above-mentioned reason, we performed calibration of the original models for the 1-year and 2-year models. Specifically, we estimated and updated the intercept of the original model ("recalibration in the large"). For the calibration plots we used the proposed methodology of Van Calster et al. (38) and the calibration plots function of Gerts et al.(39) using the "ModelGood'' package in R (40).

## Open science

Due to patient privacy we are not allowed to share the original dataset publicly. However, in order to adhere as much as possible to the open science principles, we have made available the code to generate the results presented in this manuscript on GitHub (https://github.com/riannefijten/codeErectileDysfunctionArticle). Customized data sets from the Netherlands Cancer Registry (NCR) can be made available under strict terms and conditions. All applications for data will be reviewed for their compliance with national privacy legislation as well as IKNL objectives. The dataset involved in this study can be requested via these routes and should be referred to by its reference number: K19.117.

## Results

Characteristics of the patient population in our cohort are displayed in Table 1. Our subcohort had a high prevalence of cardiovascular disease (CVD); 52% of patients had some form of cardiovascular disease. Diabetes was less prevalent, with 12% of patients suffering from it. Most patients in our subcohort received no active therapy (n=351 and 270, at 1 and 2 years respectively), and 279 & 255 patients received radical prostatectomy, 166 & 133 received EBRT and 52 & 42 patients received brachytherapy.

Table 1: Description of the patient population included in our ProZIB sub dataset

|  | 1-year cohort | 2-year cohort |
| --- | --- | --- |
| Number of patients | 848 | 670 |
| Treatments (n (%)) |  |  |
|   RP | 279 (32.9) | 225 (33.6) |
|   EBRT | 166 (19.6) | 133 (19.8) |
|   BT | 52 (6.1) | 42 (6.3) |
|   NAT | 351 (41.4) | 270 (40.3) |
| Age (mean ± stdev) | 68.4 ± 6.7 | 68.3 ± 6.6 |
| Tumor T stage (n (%)) |  |  |
|   T1 | 388 (45.7) | 319 (47.6) |
|   T2 | 351 (41.4) | 265 (39.6) |
|   T3 | 109 (12.9) | 86 (12.8) |

| Tumor N stage (n (%)) | | |
|---|---|---|
| N1 | 491 (57.9) | 395 (59.0) |
| NX | 357 (42.1) | 275 (41.0) |
| PSA at diagnosis (mean ± stdev) | 10.9 ± 13.9 | 11.4 ± 15.4 |
| ISUP grade group (n (%)) | | |
| Grade 1 | 470 (55.4) | 376 (56.1) |
| Grade 2 | 198 (23.3) | 148 (22.1) |
| Grade 3 | 75 (8.8) | 60 (9.0) |
| Grade 4 | 61 (7.2) | 49 (7.3) |
| Grade 5 | 29 (3.4) | 23 (3.4) |
| *missing* | *15 (1.8)* | *14 (2.1)* |
| Comorbidities (n (%)) * | | |
| Cardiovascular disease | 441 (52.0) | 348 (51.9) |
| Diabetes | 96 (11.3) | 75 (11.2) |
| Smoking (n (%)) | | |
| Non-smoker | 367 (43.3) | 288 (43.0) |
| Former smoker | 332 (39.2) | 264 (39.4) |
| Current smoker | 50 (5.9) | 39 (5.8) |
| *missing* | *99 (11.7)* | *79 (11.8)* |
| Alcohol use (n (%)) | | |
| No alcohol use | 83 (9.8) | 68 (10.1) |
| Previous alcohol use | 43 (5.1) | 35 (5.2) |
| Current alcohol use | 625 (73.7) | 489 (73.0) |
| *Missing* | *97 (11.4)* | *78 (11.6)* |

*\* Only CVD and diabetes are mentioned due to their significance to our outcomes.*

Preprocessing resulted in two datasets; one for each outcome time point. The dataset for the 1-year outcome consisted of 848 patients and the 2-year dataset contained 670 patients due to a higher level of non-response on the EPIC-26 questionnaire at this later time point. These datasets were also checked for non-outcome related batches, but no batch effects were found.

Furthermore, the distribution of the ED outcome of all patients in the dataset is reported in figure 2. Full ED is the most prevalent outcome at both time points: 46% and 47% could never achieve an erection at 1 and 2 years post-diagnosis respectively. The remaining 54% and 53% reported being able to have an erection to some extent (figure 2).

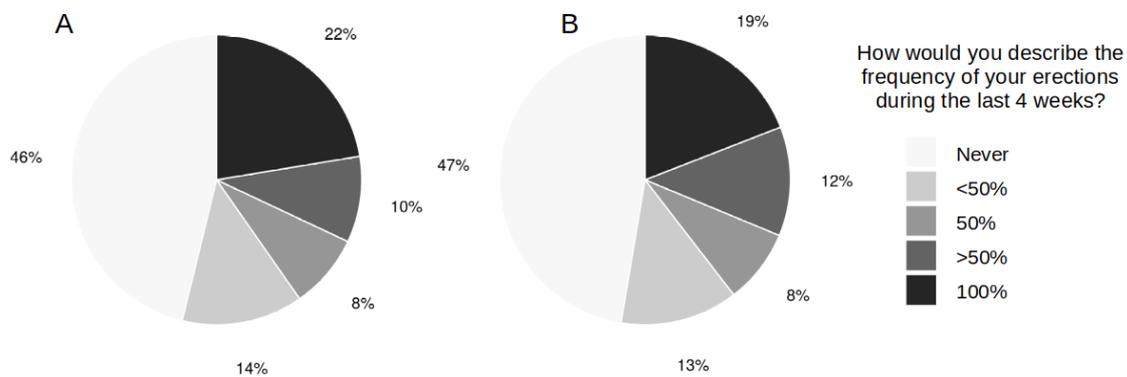

Figure 2: Distribution of the ED answers given by patients 1 year (A) and 2 years (B) after diagnosis.

Subsequently, the dataset was split up into a training and validation set and tested for differences. In the 1-year dataset, no significant differences were found between the training- and test set for age (Wilcoxon p = 0.72), tumor T stage (Wilcoxon p = 0.95), tumor N stage (Wilcoxon p = 0.09), PSA at diagnosis (Wilcoxon p = 0.46) and ISUP grade group (Wilcoxon p = 0.07). In the 2-year dataset, significant differences were found between the training and test set for tumor T and N stage (Wilcoxon p = 0.02 for both), but not for age (Wilcoxon p = 0.12), PSA at diagnosis (Wilcoxon p = 0.12), and ISUP grade group (Wilcoxon p = 0.80).

## Model training

Two models were created. The first model predicted the 1-year ED outcome based on 10 variables with a training accuracy of 77.7% with a sensitivity and specificity of 89.9% and 66.8% respectively. The selected variables included both patient-related factors such as alcohol use and clinical variables such as the tumor T stage and the ISUP grade group. The second model predicted the 2-year outcome based on 9 variables and had a training accuracy of 73.9% with a sensitivity of 86.4% and a specificity of 62.3%. Several variables overlapped between the two models, but several were different. For instance, cardiovascular disease was important to the 1-year outcome, but not the 2-year outcome.

In order to encourage reproducibility, table 2 and table 3 contain the logistic regression coefficients required to compute the model. Calibration-in-the-large is calculated but there is not much difference with the coefficient provided in these tables [for 1-year the updated intercept is 0.05485955 and for the 2-year is 0.1988651]. In these tables the univariate statistical significance of each variable independently is displayed. These results show that each of the selected variables is statistically significant except for alcohol use for the 1-year outcome and abdominal/pelvic/rectal pain for the 2-year outcome. Other variables that were not selected by the algorithm were also significant between the patients that could not get an erection and those that still could, such as the PSA at diagnosis and sexual functioning. These are shown in detail in the appendix.

Table 2: An overview of the logistic regression coefficients for the input variables in the 1-year model. In addition, the univariate FDR-corrected q-value is reported for each variable.

| Variable | Coefficients | FDR corrected q-value |
| --- | --- | --- |
| Intercept | -2.081 | NA |
| The four major treatments groups | 0.900 | 6.21E-33 |
| Pre-treatment quality of erections | 0.540 | 5.19E-23 |
| Pre-treatment frequency of erections | 0.380 | 1.6E-22 |
| ISUP grade group | -0.334 | 4.93E-22 |
| Tumor T stage | -0.091 | 1.37E-11 |
| Hormone therapy given to patients | -0.696 | 3.22E-07 |
| The presence of cardiovascular disease (CVD) | -0.209 | 0.003 |
| The presence of diabetes | -0.690 | 0.007 |
| Lack of energy | -0.227 | 0.016 |
| Alcohol use | -0.023 | 0.053 |

Table 3: The logistic regression coefficients of the input variables in the 2-year outcome prediction model. In addition, the univariate FDR-corrected q-value is reported for each variable.

| Variable | Coefficients | FDR corrected q-value |
| --- | --- | --- |
| (Intercept) | -3.751 | NA |
| Pre-treatment quality of erections | 0.633 | 6.15E-22 |
| Pre-treatment frequency of erections | 0.330 | 1.07E-20 |
| The four major treatments groups | 0.661 | 9.19E-17 |
| ISUP grade group | -0.258 | 2.48E-13 |
| Tumor T stage | -0.197 | 1.09E-08 |
| Charlson comorbidity index simplified | -0.175 | 8.83E-05 |
| Hormone therapy given to patients | -0.079 | 0.003 |
| The presence of diabetes | -0.253 | 0.003 |
| Abdominal/pelvic/rectal pain | 0.388 | 0.177 |

## Model validation

After model training, the (independent) validation set was used to validate the model and thus identify overfitting. The accuracy, sensitivity and specificity of the models on the test set are depicted in table 4 in addition to the training set values for comparison. For both time points, the sensitivity and specificity between the training and test set are similar with only a few percent difference. The overall accuracy of the 1-year model was 3.1% higher than that of the 2-year model and had a better specificity.

Table 4: The number of variables, sensitivity, specificity and overall accuracy of both models that predict reliability of the erection at 1 year and 2 years post-diagnosis.

| | No. of variables | Sensitivity (%) | Specificity (%) | Overall accuracy (%) |
|---|---|---|---|---|
| Reliability after 1 year | 10 | | | |
|    Training set | | 89.9 | 66.8 | 77.7 |
|    Test set | | 87.1 | 66.2 | 75.3 |
| Reliability after 2 years | 9 | | | |
|    Training set | | 86.4 | 62.3 | 73.9 |
|    Test set | | 87.4 | 62.1 | 73.7 |

The associated validation set ROC curves are shown in figure 3 and report AUCs of 0.84 and 0.81 for 1 year and 2 years post-diagnosis respectively.

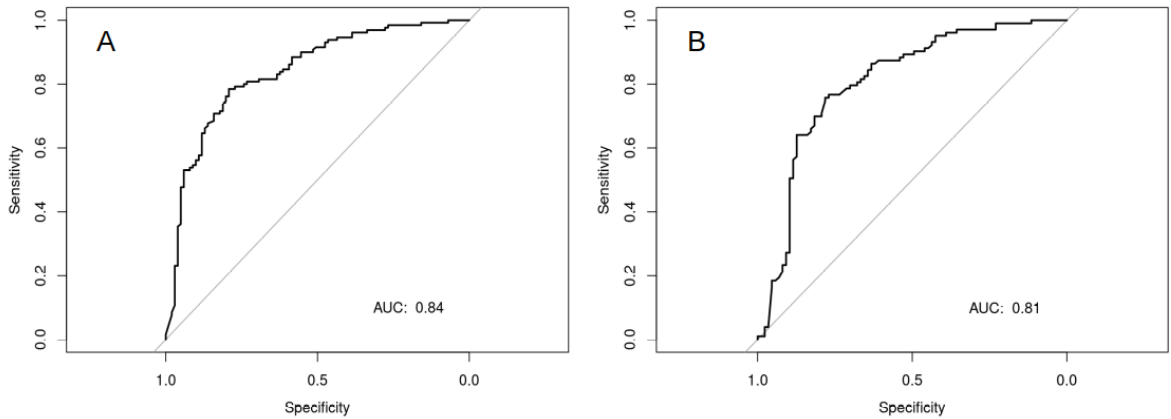

Figure 3: the ROC curves of the independent test set for the model predicting the reliability 1 year (A) and 2 years (B) post-diagnosis.

## Calibration

We generated the calibration plots for the original and the calibrated model for the 1-year and 2-year models that displayed in figures 4 and 5. The curves of the calibrated models did not show a significant difference in terms of overestimating or underestimating the risk of erectile dysfunction compared to the original models.

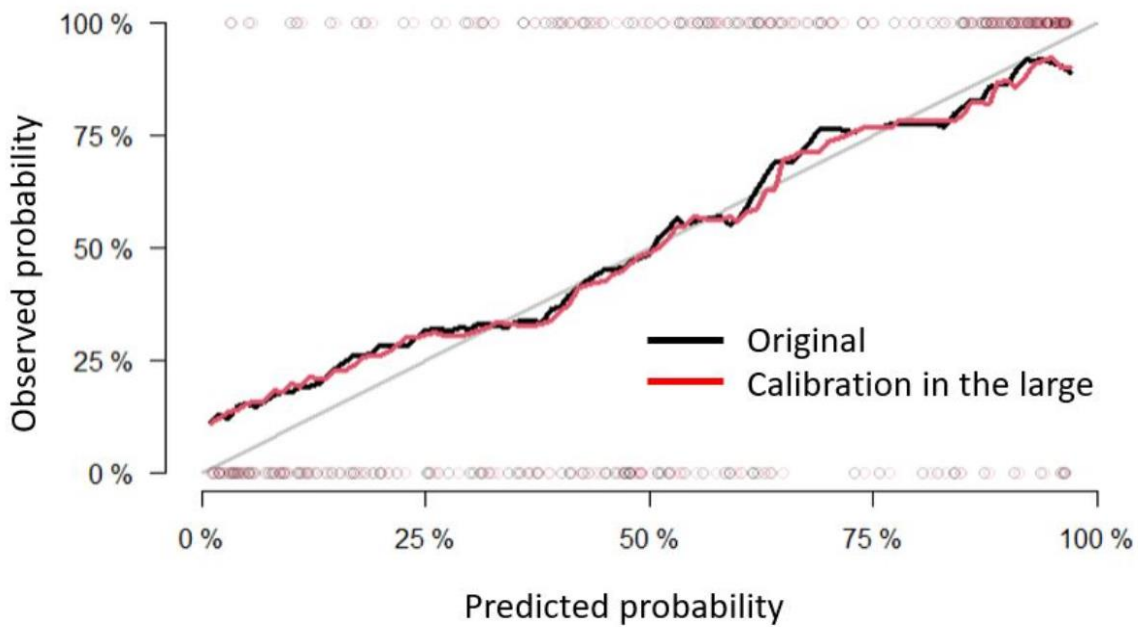

Figure 4: a calibration plot of the 1-year erectile dysfunction model.

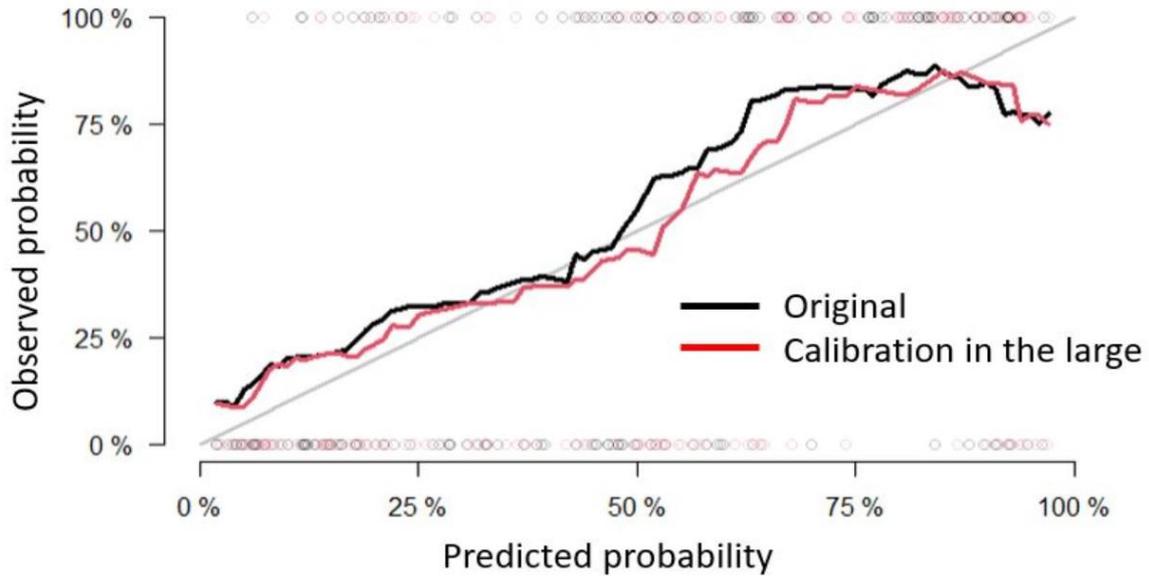

Figure 5: a calibration plot of the 2-year erectile dysfunction model.

## Nomograms

In addition to reporting the coefficients and accuracy of the models, we generated nomograms for both models. The nomogram for the 1-year model is displayed in figure 6 and the one for the 2-year model is displayed in figure 7.

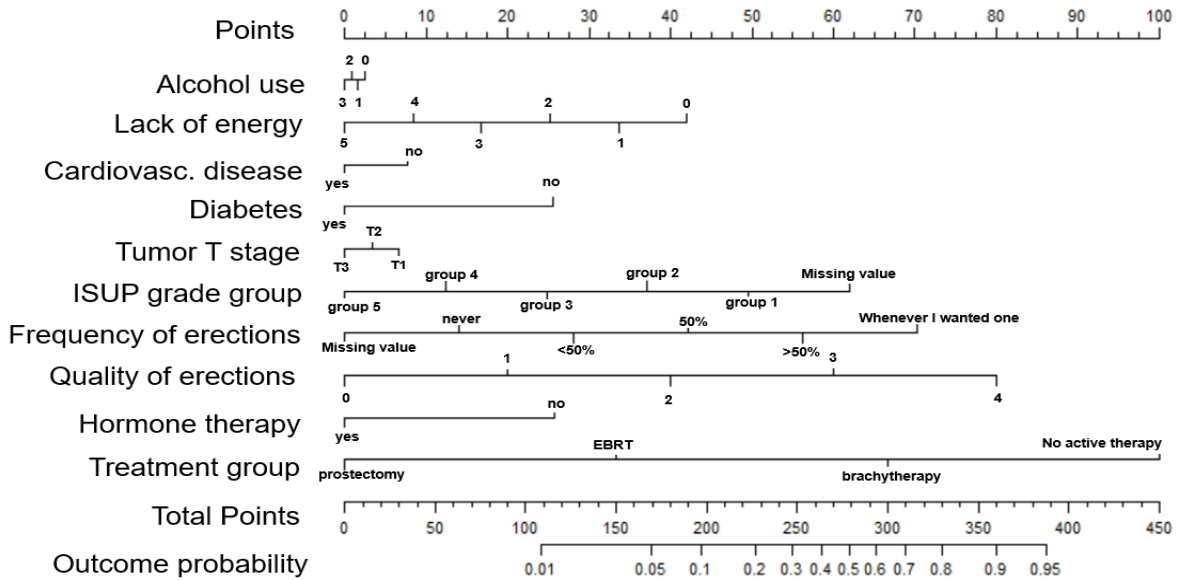

Figure 6: a nomogram of the 1-year erectile dysfunction model.

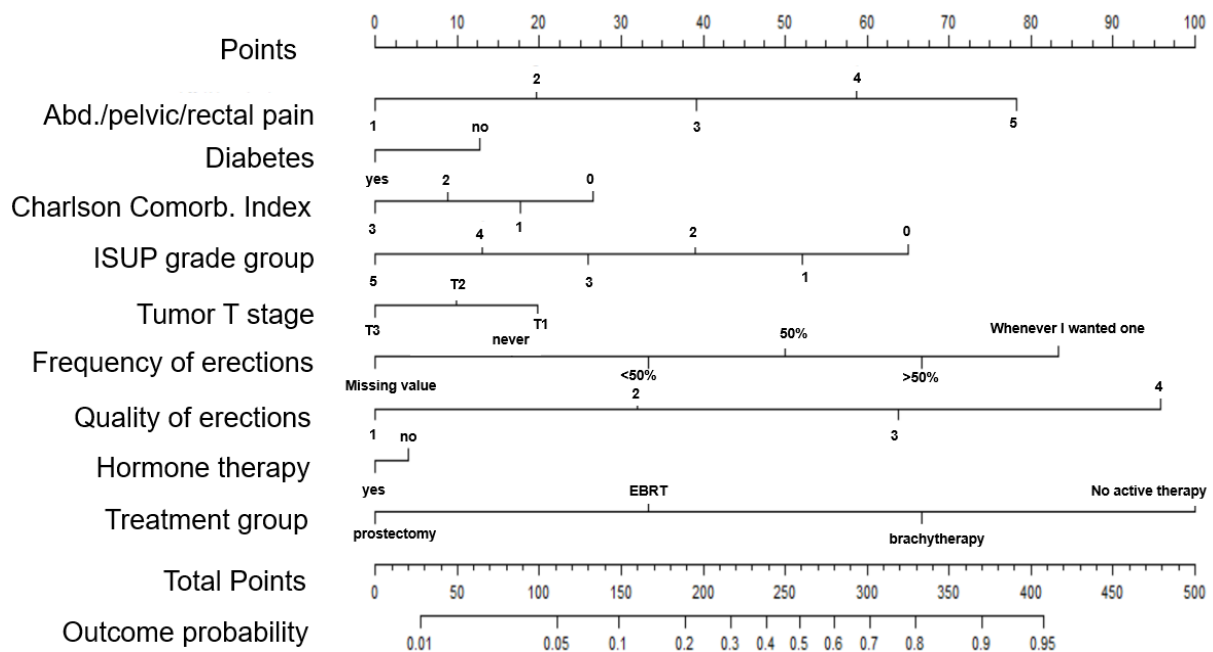

Figure 7: a nomogram of the 2-year erectile dysfunction model.

# Discussion

In this study, we developed and validated a logistic regression model (TRIPOD type 3) that predicts the frequency of ED 1 and 2 year post-diagnosis based on data available at the moment of diagnosis. These models are 75.3% and 73.7% accurate in the external validation set for the 1-year and 2-year outcomes respectively.

The variables that were able to predict the ED outcome were related to the tumor as well as patient-related factors; for example the tumor T stage and the quality of their erections at diagnosis. The two included comorbidities were also predictive of the outcomes, with cardiovascular disease predictive of the 1-year outcome and diabetes as indicative of the outcome at both time points. This is consistent with previous literature that these two diseases influence ED in patients with prostate cancer (23,24).

In both validations, the sensitivity (the accuracy for patients that could never have an erection) was substantially higher than the specificity (the accuracy for patients that could still get an erection). This is most likely a result of the heterogeneity introduced by combining patients with ED complaints ranging from mild to severe for our binary classification with logistic regression.

Despite ED being the most frequent side effect of prostate cancer treatment, there are not many suitable and clinically applicable models in scientific literature that predict the ED chance at a certain time point with data available upon diagnosis. Furthermore, there are currently no AI-based prediction models that predict ED for patients that undergo active surveillance. One study developed a web-based prediction model to predict common prostate cancer side effects such as sexual function after treatment or without active therapy, but there is no information available about the accuracy of this model (41).

Oh et al. investigated the use of genetic markers to determine the chances of developing ED after radiotherapy in 236 patients at a single institute. The data was collected from patients with at least one year follow up. They used random forest regression to build the model and internally validated the model (19). The AUC of the model is 0.65, which is lower than our models. In addition, genetic markers are most likely not available at diagnosis which could hamper its clinical applicability.
In addition, Haskins et al. predicted the chance of ED 2 years post-treatment in 776 patients who underwent prostatectomy and radiotherapy based on pretreatment conditions. Multivariate logistic regression was used and resulted in an AUC of 0.71 for prostatectomy and 0.66 for radiotherapy, which is lower than the accuracy of the models in our study. The model was not externally validated, but internally validated by bootstrapping (42).

A study similar to ours predicted ED two years after treatment based on pretreatment demographics and clinical data of 1027 patients from nine centers in the USA. This model was validated externally on 1913 patients and achieved an AUC of 0.77 for prostatectomy, 0.87 for external radiotherapy and 0.90 for brachytherapy (43). This model performs similarly well to our model, but uses different variables, such as age and BMI. In our models, age was not selected as a predictive variable to our outcome. We did not have sufficient data to include the BMI. One variable was reported by both studies, namely hormone therapy.

In addition to the statistical performance of a model, it is important that the predictors identified in our study are clinically explainable and relevant to patients (44). Our prediction models were based on a combination of clinical information, such as tumor T stage and the presence of cardiovascular disease or diabetes, and patient characteristics, such as the presence of erectile dysfunction at baseline. The most common predictors mentioned in literature are age, pretreatment potency and neurovascular bundle injury during treatment (45,46). Preservation of the neurovascular bundle is however a variable that will not be known at diagnosis and is therefore not applicable to our use-case. Alemozaffar et al (43). showed that the nerve sparing surgery technique is related to the chance of keeping an erection. Age was included in our analysis, but was not selected as predictive for ED. Pretreatment potency was defined by two variables in our dataset (the erectile frequency and quality) and was found as predictive of ED after 1 or 2 years. In addition, previous research has found a correlation between diabetes and treatment-related ED (45). This is consistent with our findings that diabetes is predictive of ED at both time points, yet was not found by Alemozaffar et al. (43).

The aim of this study was to identify patients who are at the risk of developing ED after common prostate cancer treatments as well as the degree of importance based on demographics, clinical and PROMs of patients. Results show that regardless of the type of treatment, 78% of patients experienced the disorder, completely or partially, one year after diagnosis and 81% two years after diagnosis. However, it is not clear which combination of treatments and identified factors reduce the risk of ED. In other words, does a 70-year-old patient with a history of cardiovascular disease benefit more from surgery or radiotherapy? The importance of this topic is for guiding SDM sessions from population-based to individual-based, implementing personalized medicine and deciding on the optimal treatment. Further studies are needed to identify subgroups of patients who have a lower chance of developing ED with the benefit of a certain treatment.

Despite its strengths, this study has some limitations. First, we used regression logistics in this study, which investigates the occurrence ED in a binary way. However, ED is classified into different degrees in terms of quality and frequency. Further research is needed to explore the relationship between ED and predictors with classification models such as linear regression. Second, as all the data were collected from hospitals in the Netherlands, results might differ in other regions or countries (47). For instance, different rates of ethnic diversity and different cancer detection and treatment execution in different regions could influence the applicability of our models to other populations.  We would therefore highly encourage researchers from other countries to validate our model with their own data. In addition, the data shows a clear preference in the treatment decision making for prostatectomy and watchful waiting, with much fewer patients receiving brachytherapy or EBRT. Our models could therefore potentially be less accurate for these patients. Furthermore, only two comorbidities (CVD and diabetes) were considered as risk factors for ED post treatment due to the evidence for their role in ED development in scientific literature. Therefore, other comorbidities might play a role in ED development.

In conclusion, we developed and externally validated prediction models for ED at 1 and 2 year post-diagnosis based on data from Dutch prostate cancer patients. In this study we identified tumor T stage, frequency and quality of erections, lack of energy, ISUP group, presence of CVD, diabetes, hormone therapy, and treatment group as the most important predictors of posttreatment ED. Our models perform similar or better than most other scientific literature and will be incorporated in the patient decision aid (PDA) to assist physicians and patients make better informed and evidence-based decisions about localized prostate cancer with the emphasis on QoL.